\documentclass[11pt, letterpaper]{article}
\usepackage[letterpaper]{geometry}
\usepackage{acl2012}
\usepackage{times}
\usepackage{latexsym}

\usepackage{caption}
\usepackage{algorithm}
\usepackage{algorithmicx}
\usepackage{algpseudocode}

\usepackage{multirow}
\usepackage{tikz}
\usepackage{pgfplots}
\usepackage{array}
\usepackage{amsmath}
\makeatletter

\title{Shift-Reduce Constituent Parsing with Neural Lookahead Features}

\author{Jiangming Liu \and Yue Zhang\\
	    Singapore University of Technology and Design,\\
	    8 Somapah Road, Singapore, 487372\\
	    {\tt \{jiangming\_liu, yue\_zhang\}@sutd.edu.sg}}

\date{}

\begin{document}

\maketitle

\begin{abstract}
Transition-based models can be fast and accurate for constituent parsing.
Compared with chart-based models, they leverage richer features by extracting history information from a parser stack, which spans over non-local constituents.
On the other hand, during incremental parsing, constituent information on the right hand side of the current word is not utilized, which is a relative weakness of shift-reduce parsing.
To address this limitation, we leverage a fast neural model to extract lookahead features. 
In particular, we build a bidirectional LSTM model, which leverages the full sentence information to predict the hierarchy of constituents that each word starts and ends.
The results are then passed to a strong transition-based constituent parser as lookahead features.
The resulting parser gives 1.3\% absolute improvement in WSJ and 2.3\% in CTB compared to the baseline, given the highest reported accuracies for fully-supervised parsing.
\end{abstract}

\section{Introduction}
Transition-based constituent parsers are fast and accurate, performing incremental parsing using a sequence of state transitions in linear time.
Pioneering models rely on a classifier to make local decisions, searching greedily for local transitions to build a parse tree \cite{Sagae:2005dw}.
\newcite{Zhu:2013up} use a beam search framework, which preserves linear time complexity of greedy search, while alleviating the disadvantage of error propagation.
The model gives state-of-the-art accuracies at a speed of 89 sentences per second on the standard WSJ benchmark \cite{Marcus:1993wd}.

\newcite{Zhu:2013up} exploit rich features by extracting history information from a parser stack, which spans over non-local constituents.
However, due to the incremental nature of shift-reduce parsing, the right-hand side constituents of the current word cannot be used to guide the action at each step.
Such lookahead features \cite{Tsuruoka:2011uh} correspond to the outside scores in chart parsing \cite{Goodman:1998tn}.
It has been shown highly effective for obtaining improved accuracies.

To leverage such information for improving shift-reduce parsing, we propose a novel neural model to predict the constituent hierarchy related to each word before parsing.
Our idea is inspired by the work of \newcite{Roark:2009vd} and \newcite{Zhang:2010vo}, which shows that shallow syntactic information gathered over the word sequence can be utilized for improving chart parsing speed without losing accuracies.
For example, \newcite{Roark:2009vd} predict constituent boundary information on words as a pre-processing step, and use such information to prune the chart.
Since such information is much lighter-weight compared to full parsing, it can be predicted relatively accurately using sequence labellers. 

%%%%%%%constituent feature example
\begin{figure*}
\begin{center}
\includegraphics[width=16cm,height=4.92cm]{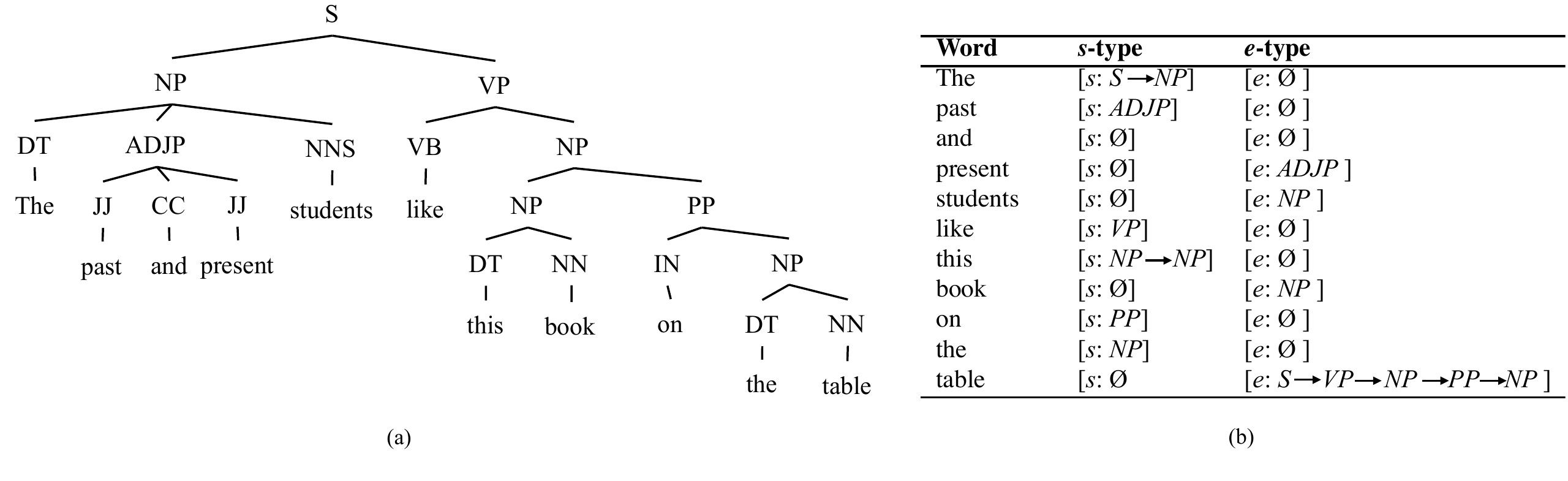}
\end{center}
\caption{\label{constituent feature example} Example constituent hierarchies on the sentence `` The past and present students like this book on the table". (a) parse tree; (b) constituent hierarchies on words.}
\end{figure*}
%%%%%%%

Different from that of \newcite{Roark:2009vd}, we collect \emph{lookahead} constituent information for \emph{shift-reduce} parsing, rather than \emph{pruning} information for \emph{chart} parsing.
In addition, our concern is in the \emph{accuracies} rather than the \emph{speed}.
Accordingly, our model should predict the constituent hierarchy on each word rather than simple boundary information.
For example, in Figure \ref{constituent feature example}(a), the constituent hierarchy that the word ``\emph{The}" starts is ``\emph{S} $\rightarrow$ \emph{NP}", and the constituent hierarchy that the word ``table" ends is ``\emph{S} $\rightarrow$ \emph{VP} $\rightarrow$ \emph{NP} $\rightarrow$ \emph{PP} $\rightarrow$ \emph{NP}".
For each word, we predict both the constituent hierarchy it starts and the constituent hierarchy it ends, using them as lookahead features.

The task is challenging in three aspects.
First, it is significantly more difficult compared to simple sequence labelling, since two sequences of constituent hierarchies must be predicted for each word in the input sequence.
Second, for high accuracies, global features from the full sentence are necessary since constituent hierarchies contain rich structural information.
Third, to retain high speed for shift-reduce parsing, lookahead feature prediction must be executed efficiently.
It is highly difficult to build such a model using manual discrete features and structured search.

Fortunately, sequential recurrent neural networks (RNNs) are remarkably effective models to encode the full input sentence.
We leverage RNNs for building our constituent hierarchy predictor.
In particular, a LSTM \cite{Hochreiter:1997fq} is used to learn global features automatically from the input words.
For each word, a second LSTM is then used to generate the constituent hierarchies greedily using features from the hidden layer of the first LSTM, in the same way as a neural language model decoder generating output sentences for machine translation \cite{Bahdanau:2014vz}.
The resulting model solves all the three challenges raised above.
For fully-supervised learning, we learn word embeddings as a part of the model parameters.

In the standard WSJ \cite{Marcus:1993wd} and CTB 5.1 tests \cite{Xue:2005gq}, our parser gives 1.3 $F_1$ and 2.3 $F_1$ improvement, respectively, over the baseline of \newcite{Zhu:2013up}, resulting in a accuracy of 91.7 $F_1$ for English and 85.5 $F_1$ for Chinese, which are the best for fully-supervised models in the literature.
We release our code, based on ZPar, at https://github.com/SUTDNLP/LookAheadConparser.

% section baseline system
\section{Baseline System}
We adopt the parser of \newcite{Zhu:2013up} for a baseline, which is based on the shift-reduce process of \newcite{Sagae:2005dw} and the beam search strategy of \newcite{zhang2011syntactic} with the global perceptron training.

%%%%%%% Deduction system
\begin{figure}[!tp]
\begin{center}
\renewcommand{\arraystretch}{0.8}
\begin{tabular}{>{\small}c>{\small}c}
Initial State & $[\phi, 0, false, 0]$\\
Final State & $[S, n, true, m: 2n<= m <= 4n]$\\
\\
\multicolumn{2}{>{\small}c}{Induction Rules:} \\
\textsc{Shift} & {\Large$\frac{[S,i, false, k]}{[S|w, i+1, false, k+1]}$} \\
\\
\textsc{Reduce-l/r-x} & {\Large$\frac{[S|s_1s_0, i, false, k]}{[S|X,i,false, k+1]}$} \\
\\
\textsc{Unary-x} & {\Large$\frac{[S|s_0, i, false, k]}{[S|X,i,false,k+1]}$} \\
\\
\textsc{Finish} & {\Large$\frac{[S, n, false, k]}{[S,n,true,k+1]}$} \\
\\
\textsc{Idle} & {\Large$\frac{[S, n, true, k]}{[S,n,true,k+1]}$} \\
\end{tabular}
\end{center}
\caption{\label{deduction system} Deduction system for the baseline shift-reduce parsing process.}
\end{figure}
%%%%%%%

%% shift-reduce constituency parsing
\subsection{The Shift-Reduce System}
Shift-reduce parsers process an input sentence incrementally from left to right.
A stack is used to maintain partial phrase-structures, while the incoming words are ordered in a buffer.
At each step, a transition action is applied to consume an input word or construct a new phrase-structure.
The set of transition actions are 
\begin{itemize}
\item \textsc{Shift}: pop the front word off the buffer, and push it onto the stack. 
\item \textsc{Reduce-l/r-x}: pop the top two constituents off the stack, combine them into a new constituent with label X, and push the new constituent onto the stack. 
\item \textsc{Unary-x}: pop the top constituent off the stack, raise it to a new constituent X, and push the new constituent onto the stack. 
\item \textsc{Finish}: pop the root node off the stack and end parsing. 
\item \textsc{Idle}: no-effect action on a completed state without changing items on the stack or buffer.
\end{itemize} 
The deduction system for the process is shown in Figure \ref{deduction system}, where a state is represented as [\emph{stack}, \emph{buffer front index}, \emph{completion mark}, \emph{action index}], and $n$ is the number of words in the input.
For example, given the sentence ``They like apples", the action sequence ``\textsc{Shift}, \textsc{Shift}, \textsc{Shift}, \textsc{Reduce-L-VP}, \textsc{Reduce-R-S}" gives its syntax ``(S They (VP like apples) )".

%% section training
\subsection{Search and Training}
Beam-search is used for decoding with the $k$ best state items at each step being kept in the agenda. During initialization, the agenda contains only the initial state [$\phi, 0, false, 0$]. At each step, each state in the agenda is popped and expanded by applying all valid transition actions, and the top $k$ resulting states are put back onto the agenda \cite{Zhu:2013up}. The process repeats until the agenda is empty, and the best completed state is taken as output.

The score of a state is the total score of the transition actions that have been applied to build it:
\begin{equation} 
C(\alpha)=\sum^{N}_{i=1}\Phi(\alpha_i)\cdot \vec{\theta}
\end{equation}
Here $\Phi(\alpha_i)$ represents the feature vector for the $i$th action $\alpha_i$ in the state item $\alpha$. $N$ is the total number of actions in $\alpha$.

The model parameter set $\vec{\theta}$ is trained online using the averaged perceptron algorithm with the early-update strategy \cite{Collins:2004gi}.

%%%%%%%multi-labelling parameter
\begin{table}[!tp]
\begin{center}
\renewcommand{\arraystretch}{1.0}
\begin{tabular}{>{\small}l>{\small}l}
\hline
\bf Description & \bf Templates \\
\hline
\hline
\textsc{Unigram}
& $s_0tc,s_0wc,s_1tc,s_1wc,s_2tc$ \\
& $s_2wc,s_3tc,s_3wc,q_0wt,q_1wt$ \\
& $q_2wt,q_3wt,s_0lwc,s_0rwc$ \\
& $s_0uwc,s_1lwc,s_1rwc,s_1uwc$ \\
\textsc{Bigram}
& $s_0ws_1w,s_0ws_1c,s_0cs_1w,s_0cs_1c$ \\
& $s_0wq_0w,s_0wq_0t,s_0cq_0w,s_0cq_0t$ \\
& $q_0wq_1w,q_0wq_1t,q_0tq_1w,q_0tq_1t$ \\
& $s_1wq_0w,s_1wq_0t,s_1cq_0w,s_1cq_0t$ \\
\textsc{Trigram}
& $s_0cs_1cs_2c,s_0ws_1cs_2c,s_0cs_1wq_0t$ \\
& $s_0cs_1cs_2w,s_0cs_1cq_0t,s_0ws_1cq_0t$ \\
& $s_0cs_1wq_0t,s_0cs_1cq_0w$ \\
\hline
Extended
& $s_0llwc,s_0lrwc,s_0luwc$ \\
& $s_0rlwc,s_0rrwc,s_0ruwc$ \\
& $s_0ulwc,s_0urwc,s_0uuwc$ \\
& $s_1llwc,s_1lrwc,s_1luwc$ \\
& $s_1rlwc,s_1rrwc,s_1ruwc$ \\
\hline
\end{tabular}
\end{center}
\caption{\label{baseline feature} Baseline feature templates, where $s_i$ represents the $i$th item on the top of the stack and $q_i$ denotes the $i$th item in the front of the buffer. The symbol $w$ denotes the lexical head of an item; the symbol $c$ denotes the constituent label of an item; the symbol $t$ is the POS of a lexical head; $s_ill$ denotes the left child of $s_i$'s left child. Other notations can be explained in a similar way.}
\end{table}
%%%%%%%

%%baseline features
\subsection{Baseline Features}
Our baseline features are taken from \newcite{Zhu:2013up}.
As shown in Table \ref{baseline feature}, they include the \textsc{Unigram}, \textsc{Bigram}, \textsc{Trigram} features of \newcite{Zhang:2009ix} and the extended features of \newcite{Zhu:2013up}. 

%constituent features
\section{Global Lookahead Features}
The baseline features suffer two limitations, as mentioned in the introduction. 
First, they are relatively local to the state, considering only the neighbouring nodes of $s_0$ (top of stack) and $q_0$ (front of buffer).
Second, they do not consider lookahead information beyond $s_3$, or the syntactic structure of the buffer and sequence. 
We use a LSTM to capture full sentential information in linear time, representing such global information into the baseline parser as a constituent hierarchy for each word.
\emph{Lookahead features} are extracted from the constituent hierarchy to provide top-down guidance for bottom-up parsing.
%%%%%%%multi-labelling parameter
\begin{table}[!tp]
\begin{center}
\renewcommand{\arraystretch}{0.8}
\begin{tabular}{>{\small}l}
\hline
\bf Templates \\
\hline
\hline
$s_0c_gs, s_0c_ge, s_1c_gs, s_1c_ge$ \\
$q_0c_gs, q_0c_ge, q_1c_gs, q_1c_ge$ \\
\hline
\end{tabular}
\end{center}
\caption{\label{lookahead feature template} Lookahead feature templates, where $s_i$ represents the $i$th item on the top of the stack and $q_i$ denotes the $i$th item in the front end of the buffer. The symbol $c_gs$ and $c_ge$ denote the next level constituent in the $s$-type hierarchy and $e$-type hierarchy, respectively.}
\end{table}
%%%%%%%
%%%%%%%constituent feature example
\begin{figure*}
\begin{center}
\includegraphics[width=16cm,height=4.92cm]{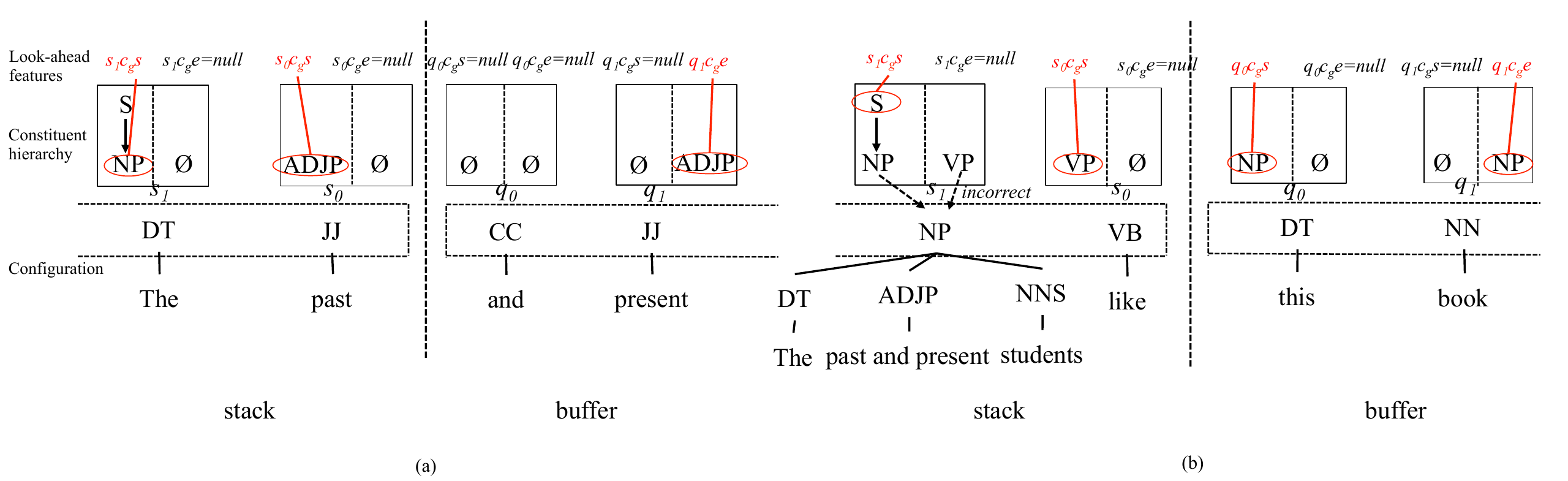}
\end{center}
\caption{\label{lookahead features extraction} Two intermediate states of parsing on the sentence ``The past and present students like this book on the table". Each item on the stack or buffer has two constituent hierarchies: s-type (left) and e-type (right), respectively, in the corresponding box. Noted that the e-type constituent hierarchy of the word ``students" are incorrectly predicted and used as soft features in our model. }
\end{figure*}
%%%%%%% 
\subsection{Constituent Hierarchy}
In a constituency tree, each word can start or end a constituent hierarchy.
As shown in Figure \ref{constituent feature example}, the word ``\emph{The}" starts a constituent hierarchy ``\emph{S} $\rightarrow$ \emph{NP}". In particular, it starts a constituent \emph{S} in the top level and then a following constituent \emph{NP}.
The word ``\emph{table}" ends a constituent hierarchy ``\emph{S} $\rightarrow$ \emph{VP} $\rightarrow$ \emph{NP} $\rightarrow$ \emph{PP} $\rightarrow$ \emph{NP}". In particular, it ends a constituent \emph{S} in the top level, and then a \emph{VP} (starting from the word ``\emph{like}"), an \emph{NP} (starting from the noun phrase ``\emph{this book}"), a \emph{PP} (starting from the word ``\emph{in}"), and finally an \emph{NP} (starting from the word ``\emph{the}").
The extraction of constituent hierarchies for each word is based on \emph{unbinarized} grammars, reflecting the start and end in \emph{unbinarized} trees.
The constituent hierarchy is \emph{empty} (denoted as $\phi$) if the corresponding words does not start or end a constituent.
The constituent hierarchies are added into the shift-reduce parser as soft features (section 3.2).
%If a word not starts or ends a constituent, the constituent hierarchy will be empty denoted as $\phi$.

Formally, a constituent hierarchy is defined as
\begin{equation*}
[type: c_1 \rightarrow c_2 \rightarrow ... \rightarrow c_z],
\end{equation*}
where $c$ is a constituent label (e.g. \emph{NP}), ``$\rightarrow$" represents the top-down hierarchy, and $type$ can be $s$ or $e$, denoting that the current word starts or ends the constituent hierarchy, respectively, as shown in Figure \ref{constituent feature example}.
Compared with full parsing, the constituent hierarchies associated with each word have no forced structural dependencies between each other, and therefore can be modelled more easily, for each word individually.
Serving as soft lookahead features rather than hard constraints, their inter-dependencies are not crucial for the main parser.

\subsection{Lookahead Features}
The lookahead feature templates are defined in Table \ref{lookahead feature template}.
In order to ensure parsing efficiency, only simple feature templates are taken into consideration.
The lookahead features of a state are instantiated for the top two items on the stack (i.e. $s_0$ and $s_1$) and buffer (i.e. $q_0$ and $q_1$).
The new function $\Phi'$ is defined to output the lookahead features vector.
The score of a state in our model is simple extended form the Formula (1):
\begin{equation*} 
C(\alpha)=\sum^{N}_{i=1}\Phi(\alpha_i)\cdot \vec{\theta} + \Phi'(\alpha_i)\cdot \vec{\theta'} 
\end{equation*}
For each word, the lookahead feature represents the next level constituent in the top-down hierarchy, which can guide bottom-up parsing.

For example, Figure \ref{lookahead features extraction} shows two intermediate states during parsing.
In Figure \ref{lookahead features extraction}(a), the $s$-type and $e$-type lookahead features of $s_1$ (i.e. the word ``\emph{The}") are extracted from the constituent hierarchy in the bottom level, namely \emph{NP} and \textsc{Null}, respectively.
On the other hand, in Figure \ref{lookahead features extraction}(b), The $s$-type lookahead feature of $s_1$ is extracted from the $s$-type constituent hierarchy of same word ``\emph{The}", but is \emph{S} based on current hierarchical level. The $e$-type lookahead feature, on the other hand, is extracted from the $e$-type constituent hierarchy of end word ``\emph{students}" of the \emph{VP} constituent, which is \textsc{Null} in the next level.
Lookahead features for items on the buffer are extracted in the same way.

The lookahead features are useful for guiding shift-reduce decisions given the current state.
For example, given the intermediate state in Figure \ref{lookahead features extraction}(a), $s_0$ has a $s$-type lookahead feature \emph{ADJP}, and $q_1$ in the buffer has $e$-type lookahead feature \emph{ADJP}.
This indicates that the two items are likely reduced into the same constituent.
Further, $s_0$ cannot end a constituent because of the empty $e$-type constituent hierarchy. As a result, the final shift-reduce parser will assign more possibility to \textsc{Shift} decision.

\section{Constituent Hierarchy Prediction}
We propose a novel neural model for constituent hierarchy prediction.
Inspired by the encoder-decoder framework for neural machine translation \cite{Bahdanau:2014vz,Cho:2014uo}, we use a LSTM to capture full sentence features, and another LSTM to generate the constituent hierarchies for each word.
Compared with a CRF-based sequence labelling model \cite{Roark:2009vd}, the proposed model has three advantages.
First, the global features can be automatically represented.
Second, it can avoid the exponentially large number of labels if constituent hierarchies are treated as unique labels.
Third, the model size is relatively small, and does not have a large effect on the final parser model.

As shown in Figure \ref{structure}, the neural network consists of three main layers, namely the \emph{input layer}, the \emph{encoder layer} and the \emph{decoder layer}.
The input layer represents each word using its characters and token information; the encoder hidden layer uses a bidirectional recurrent neural network structure to learn global features from the sentence; and the decoder layer predicts constituent hierarchies according to the encoder layer features, by using attention mechanism \cite{Bahdanau:2014vz} to softly compute the contribution of each hidden unit of the encoder.

\subsection{Input Layer}
The input layer generates a dense vector representation of each input word.
We use character embeddings to alleviate OOV problems in word embeddings \cite{Ballesteros:2015wn,santos2014learning,Kim:2015vh}, concatenating a character-embedding representation of a word to its word embedding.
Formally, the input representation $x_i$ of the word $w_i$ is computed by:
\begin{equation*}
\begin{split}
x_i&=x_{w_i}\oplus c_{i\_att} \\
c_{i\_att}&=\sum_{j}\alpha_{ij}c_{ij},
\end{split}
\end{equation*}
where $x_{w_i}$ is a word embedding vector of the word $w_i$ according to a embedding lookup table, $c_{i\_att}$ is a character embedding form of the word $w_i$, $c_{ij}$ is the $j$th character in $w_i$, and $\alpha_{ij}$ is the contribution of the character $c_{ij}$ to $c_{i\_att}$, which is computed by:
\begin{equation*}
\begin{split} 
\alpha_{ij} &= \frac{e^{f(x_{w_i}, c_{ij})}}{\sum_{k}e^{f(x_{w_i}, c_{ik})}}
\end{split}
\end{equation*}
$f$ is a non-linear transformation function based on \emph{tanh} function.

\subsection{Encoder Layer}
The encoder first uses a window strategy to represent input nodes with their corresponding local context nodes.
Formally, a windowed word representation takes the form
\begin{equation*}
\begin{split}
x'_i & = [x_{i-win}; ...; x_i; ... ; x_{i+win}].\\
\end{split}
\end{equation*}
Second, the encoder scans the input sentence and generates hidden units for each input word using a recurrent neural network (RNN), which represents features of the word from the global sequence.
Formally, given the windowed input nodes $x'_1$, $x'_2$, ..., $x'_n$ for the sentence $w_1$, $w_2$, ..., $w_n$, the RNN layer calculates a hidden node sequence $h_1$, $h_2$, ..., $h_n$.
%%%%%%%sequence labelling model
\begin{figure}
\begin{center}
\includegraphics[width=8cm,height=10.2cm]{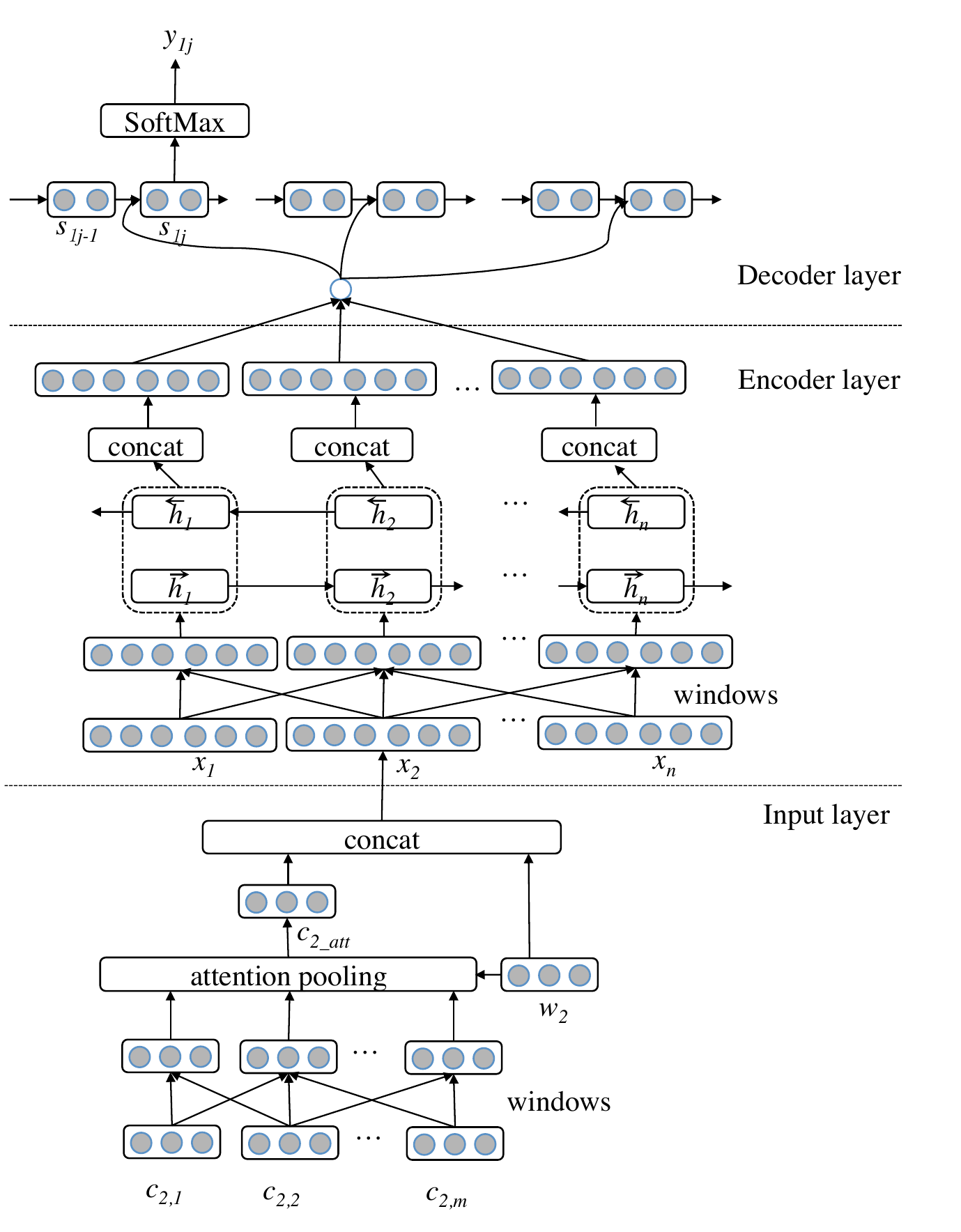}
\end{center}
\caption{\label{structure}  Structure of the constituent hierarchy prediction model. $c_{ij}$ denotes the $j$th character of the word $w_i$; $\overrightarrow{h_i}$ denotes the left-to-right encoder hidden units; $\overleftarrow{h_i}$ denotes the right-to-left encoder hidden units; $s$ denotes the decoder hidden unit; and $y_{ij}$ is the $j$th label of the word $w_i$.}
\end{figure}
%%%%%%%

Long Short-Term Memory (LSTM) mitigates the vanishing gradient problem in RNN training, by introducing gates (i.e. input $i$, forget $f$ and output $o$) and a cell memory vector $c$.
We use the variation of \newcite{Graves:2008wk}.  
Formally, the values in the LSTM hidden layers are computed as follows:
\begin{equation*}
\begin{split}
i_i &= \sigma(W_1x'_i + W_2h_{i-1} + W_3\odot c_{i-1} + b_1) \\
f_i &= 1-i_i \\
\tilde{c}_i &= tanh(W_4x'_i + W_5h_{i-1} + b_2) \\
c_i &= f_i\odot c_{i-1}+i_i\odot \tilde{c}_i \\
o_i &= \sigma(W_6x'_i + W_7h_{i-1} + W_8\odot c_{i} + b_3) \\
h_i &= o_i\odot tanh(c_i),
\end{split}
\end{equation*}
where $\odot$ is pair-wise multiplication.
Further, in order to collect features for $x_i$ from both $x'_1$, .., $x'_{i-1}$ and $x'_{i+1}$, ... $x'_n$, we use a bidirectional variation \cite{Schuster:1997fa,Graves:2013cp}.
As shown in Figure \ref{structure}, the hidden units are generated by concatenating the corresponding hidden layers of a left-to-right LSTM $\overrightarrow{h_i}$ and a right-to-left LSTM $\overleftarrow{h_i}$, where $h_i = \overrightarrow{h_i} \oplus \overleftarrow{h_i}$ for each word $w_i$.

\subsection{Decoder Layer}
The decoder hidden layer uses two different LSTMs to generate the $s$-type and $e$-type sequences of constituent labels from each encoder hidden output, respectively, as shown in Figure \ref{structure}.
Each constituent hierarchy is generated bottom-up recurrently.
In particular, a sequence of state vectors is generated recurrently, with each state yielding a output constituent label.
The process starts with a $\vec{0}$ state vector and ends when a \textsc{Null} constituent is generated.
The recurrent state transition process is achieved using LSTM model with the hidden vectors of the encoder layer being used for context features.

Formally, for the word $w_i$, the value of the $j$th state unit $s_{ij}$ of the LSTM is computed by:
\begin{equation*}
s_{ij} = f(s_{ij-1}, c_{ij}, h_i) \footnote{Here, different from typical MT model \cite{Bahdanau:2014vz}, the chain is predicted sequentially in a feed-forward way with no feedback of the prediction made. We found that this fast alternative gives similar results},
\end{equation*}
where the context $c_{ij}$ is computed by:
\begin{equation*}
\begin{split}
c_{ij} &=\sum_{k}\beta_{ijk}h_k \\
\beta_{ijk} &= \frac{e^{f(s_{ij-1}, h_{k})}}{\sum_{k'}e^{f(s_{ij-1}, h_{k'})}}
\end{split}
\end{equation*}
Here $h_k$ refers to the encoder hidden vector for $w_k$. The weight of contribution $\beta_{ijk}$ are computed under attention mechanism \cite{Bahdanau:2014vz}. The start state $s_{i,-1}=\vec{0}$.

The constituent labels are generated from each state unit $s_{ij}$, where each constituent label $y_{ij}$ is the output of a \textsc{SoftMax} function.
\begin{equation*}
p(y_{ij} = l) = \frac{e^{s_{ij}^{\top} W_l}}{\sum_{k}e^{s_{ij}^{\top} W_k}}
\end{equation*}
$y_{ij}=l$ denotes that the $j$th label of the $i$th word is $l (l \in L$).

As shown in Figure \ref{structure}, the \textsc{SoftMax} functions are applied to the state units of the decoder, generating hierarchical labels bottom-up, until the default label \textsc{Null} is predicted.

\subsection{Training}
We use two separate models to assign the $s$-type and $e$-type labels, respectively.
For training each constituent hierarchy predictor, we minimize the following training objective:
\begin{equation*}
\begin{split}
L(\theta) &= -\sum_{i}^{T}\sum_{j}^{Z_i}log~p_{ijo}+\frac{\lambda}{2}||\theta||^2,\\
\end{split}
\end{equation*}
where $T$ is the the size of the sentence, $Z_i$ is the depth of the constituent hierarchy of the word $w_i$, and $p_{ijo}$ stands for $p(y_{ij}=o)$, which is given by the \textsc{SoftMax} function, and $o$ is the golden label.

We apply back-propagation, using momentum stochastic gradient descent \cite{Sutskever:2013uw} with a learning rate of $\eta = 0.01$ for optimization and regularization parameter $\lambda = 10^{-6}$. 

%experiments
\section{Experiments}
%%data
\subsection{Experiment Settings}
English data come from the Wall Street Journal (WSJ) corpus of the Penn Treebank \cite{Marcus:1993wd}.
We use sections 2-21 for training, section 24 for system development, and section 23 for final performance evaluation.
Chinese data come from the version 5.1 of the Penn Chinese Treebank (CTB) \cite{Xue:2005gq}.
We use articles 001- 270 and 440-1151 for training, articles 301-325 for system development, and articles 271-300 were used for final performance evaluation.
For both English and Chinese data, we adopt ZPar\footnote{https://github.com/SUTDNLP/ZPar} for POS tagging, and use ten-fold jackknifing to assign auto POS tags to the training data.
In addition, we use ten-fold jackknifing to assign auto constituent hierarchies to the training data.

We use $F_1$ score to evaluate the constituent hierarchy prediction.
For example, the constituent prediction is ``\emph{S} $\rightarrow$ \emph{S} $\rightarrow$ \emph{VP} $\rightarrow$ \emph{NP}" and the golden is ``\emph{S} $\rightarrow$ \emph{NP} $\rightarrow$ \emph{NP}".
The evaluation starts from the bottom to the top, and the precision is 2/4 = 0.5, the recall is 2/3 = 0.66 and the $F1$ score is 0.57.
The metric evaluates the precision and recall of each label in the constituent hierarchy.
A label is counted as correct if and only if it occurs in the correct position index.
We use \textsc{Evalb} to evaluate parsing performance, including labelled precision ($LP$), labelled recall ($LR$), and bracketing $F_1$.\footnote{http://nlp.cs.nyu.edu/evalb}
%%multi-labeling settings
\subsection{Model Settings}
For training the constituent hierarchy prediction model, gold constituent labels are derived from labelled constituency trees in the training data.
The hyper-parameters are chosen according to development tests, and the values are shown in Table \ref{hyper-parameters}.
%%%%%%%multi-labelling parameter
\begin{table}[!tp]
\begin{center}
\renewcommand{\arraystretch}{0.8}
\begin{tabular}{|>{\small}l|>{\small}c|}
\hline
\bf hyper-parameters & \bf value \\
\hline
\hline
Word embedding size & 50 \\
Word windows & 2 \\
Character embedding size & 30 \\
Character windows & 2 \\
\hline
LSTM hidden layer size & 100 \\
Character hidden layer size & 60 \\
\hline
\end{tabular}
\end{center}
\caption{\label{hyper-parameters} Hyper-parameter settings}
\end{table}
%%%%%%%

%%%%%%%multi-labelling result
\begin{table}[!tp]
\begin{center}
\renewcommand{\arraystretch}{0.8}
\begin{tabular}{>{\small}l|>{\small}c|>{\small}c|>{\small}c}
\hline
& $s$-type & $e$-type & parser \\
\hline
\hline
1-layer & 93.39 & 81.50 & 90.43\\
2-layer & 93.76 & 83.37 & 90.72\\
3-layer & 93.84 & 83.42 & 90.80\\
%1-layer & 93.81 & 87.87 & 90.43\\
%2-layer & 94.18 & 89.06 & 90.72\\
%3-layer & 94.27 & 89.07 & 90.80\\
\hline
\end{tabular}
\end{center}
\caption{\label{label results}  Performance of the constituent hierarchy predictor and the corresponding parser on the WSJ dev dataset. $n$-layer denotes a LSTM model with $n$ hidden layers.}
\end{table}
%%%%%%% 

%%%%%%%multi-labelling result
\begin{table}[!tp]
\begin{center}
\renewcommand{\arraystretch}{0.8}
\begin{tabular}{>{\small}l|>{\small}c|>{\small}c|>{\small}c}
\hline
& $s$-type & $e$-type & parser \\
\hline
\hline
\emph{all} & 93.76 & 83.37 & 90.72 \\
\emph{all w/o wins} & 93.62 & 83.34 & 90.58 \\
\emph{all w/o chars} & 93.51 & 83.21 & 90.33 \\
\emph{all w/o chars \& wins} & 93.12 & 82.36 & 89.18 \\
%\emph{all} & 94.18 & 89.06 & 90.72 \\
%\emph{all w/o chars} & 94.01 & 88.83 & 90.33 \\
%\emph{all w/o chars \& wins} & 93.52 & 88.29 & 89.18 \\
\hline
\end{tabular}
\end{center}
\caption{\label{ablation}  Performance of the constituent hierarchy predictor and the corresponding parser on the WSJ dev dataset. \emph{all} denotes the proposed model without ablation.}
\end{table}
%%%%%%% 

For the shift-reduce constituency parser, we set the beam size to 16 for both training and decoding, which achieves a good tradeoff between efficiency and accuracy \cite{Zhu:2013up}.
The optimal training iteration number is determined on the development sets.
%% results on multi-labelling
\subsection{Results of Constituent Hierarchy Prediction}
Table \ref{label results} shows the results of constituent hierarchy prediction, where word and character embeddings are randomly initialized, and fine-tuned during training.
The third column shows the development parsing accuracies when the labels are used for lookahead features.
As Table \ref{label results} shows, when the number of hidden layer increase, both $s$-type and $e$-type constituent hierarchy prediction improve.
The accuracies of $e$-type prediction is relatively lower due to right-branching in the treebank, which makes $e$-type hierarchies longer than $s$-type hierarchies.
In addition, a 3-layer LSTM does not give significantly improvements compared to 2-layer LSTM.
For tradeoff between efficiency and accuracy, we choose the 2-layer LSTM as our constituent hierarchy predictor.

Table \ref{ablation} shows ablation results of constituent hierarchy prediction given by different reduced architectures, which include an architecture without character embeddings and an architecture with neither character embeddings nor input windows.
We find that the original architecture achieves the highest performance on constituent hierarchy prediction, compared to the two baselines.
The baseline only without the character embeddings has relatively small influence on constituent hierarchy prediction.
On the other hand, the baseline only without the input word windows has relatively smaller influence on constituent hierarchy prediction. 
Nevertheless, both of these two ablation architectures lead to much lower parsing accuracies.
The baseline removing both the character embeddings and the input word windows has relatively low accuracies.

%%%%%%%english final
\begin{table}[!tp]
\begin{center}
\renewcommand{\arraystretch}{0.8}
\begin{tabular}{>{\small}l>{\small}c>{\small}c>{\small}c}
\hline
Parser & LR & LP & F$_1$ \\
\hline
\hline
\multicolumn{4}{>{\small}l}{Fully-supervised}\\
\hline
\newcite{Ratnaparkhi:1997uj} & 86.3 & 87.5 & 86.9 \\
\newcite{Charniak:2000vi} & 89.5 & 89.9 & 89.5 \\
\newcite{collins2003head} & 88.1 & 88.3 & 88.2 \\
\newcite{Sagae:2005dw}$\dagger$ & 86.1 & 86.0 & 86.0 \\
\newcite{Sagae:2006jn}$\dagger$ & 87.8 & 88.1 & 87.9 \\
\newcite{Petrov:2007uu} & 90.1 & 90.2 & 90.1 \\
\newcite{Carreras:2008gt} & 90.7 & 91.4 & 91.1\\
\newcite{Shindo:2012wk} & N/A & N/A & 91.1 \\
\newcite{Zhu:2013up}$\dagger$ & 90.2 & 90.7 & 90.4 \\
\newcite{Socher:2013tn}* & N/A & N/A & 90.4\\
\newcite{Vinyals:2015uh}* & N/A & N/A& 88.3 \\
\bf This work & \bf 91.3 & \bf 92.1 & \bf 91.7 \\
\hline
\hline
\multicolumn{4}{>{\small}l}{Ensemble}\\
\hline
\newcite{Shindo:2012wk} & N/A & N/A & 92.4 \\
\newcite{Vinyals:2015uh}* & N/A & N/A& 90.5 \\
\hline
\hline
\multicolumn{4}{>{\small}l}{Rerank}\\
\hline
\newcite{Charniak:2005vz} & 91.2 & 91.8 & 91.5 \\
\newcite{Huang:2008ti} & 92.2 & 91.2 & 91.7 \\
\hline
\hline
\multicolumn{4}{>{\small}l}{Semi-supervised}\\
\hline
\newcite{McClosky:2006iw} & 92.1 & 92.5 & 92.3 \\
\newcite{Huang:2009vg} & 91.1 & 91.6 & 91.3 \\
\newcite{Huang:2010tn} & 91.4 & 91.8 & 91.6 \\
\newcite{Zhu:2013up}$\dagger$ & 91.1 & 91.5 & 91.3 \\
\newcite{Durrett:2015wy}* & N/A & N/A & 91.1 \\
\newcite{Dyer:2016uz}*$\dagger$ & N/A& N/A & 92.4 \\
\hline
\end{tabular}
\end{center}
\caption{\label{english final} Comparison of related work on the WSJ test set. * denotes neural parsing; $\dagger$ denotes shift-reduce framework.}
\end{table}
%%%%%%%chinese final
\begin{table}[!tp]
\begin{center}
\renewcommand{\arraystretch}{0.8}
\begin{tabular}{>{\small}l>{\small}c>{\small}c>{\small}c}
\hline
Parser & LR & LP & F$_1$ \\
\hline
\hline
\multicolumn{4}{>{\small}l}{Fully-supervised}\\
\hline
\newcite{Charniak:2000vi} & 79.6 & 82.1 & 80.8 \\
\newcite{bikel2004parameter} & 79.3 & 82.0 & 80.6 \\
\newcite{Petrov:2007uu} & 81.9 & 84.8 & 83.3 \\
\newcite{Zhu:2013up}$\dagger$ & 82.1 & 84.3 & 83.2 \\
\newcite{Wang:2015gk}$\ddagger$ & N/A & N/A & 83.2 \\
\bf This work & \bf 85.2 & \bf 85.9 & \bf 85.5 \\
\hline
\hline
\multicolumn{4}{>{\small}l}{Rerank}\\
\hline
\newcite{Charniak:2005vz} & 80.8 & 83.8 & 82.3 \\
\hline
\hline
\multicolumn{4}{>{\small}l}{Semi-supervised}\\
\hline
\newcite{Zhu:2013up}$\dagger$ & 84.4 & 86.8 & 85.6 \\
Wand and Xue \shortcite{Wang:2014bj}$\ddagger$ & N/A & N/A & 86.3 \\
Wang et al. \shortcite{Wang:2015gk}$\ddagger$ & N/A & N/A & 86.6 \\
Dyer et al. \shortcite{Dyer:2016uz}*$\dagger$ & N/A& N/A & 82.7 \\
\hline
\end{tabular}
\end{center}
\caption{\label{chinese final} Comparison of related work on the CTB5.1 test set. * denotes neural parsing; $\dagger$ denotes shift-reduce framework; $\ddagger$ denotes joint POS tagging and parsing.}
\end{table}
%%final results
\subsection{Final Results}
For English, we compare the final results with previous related work on the WSJ test sets.
As shown in Table \ref{english final}\footnote{We treat the methods as semi-supervised if they use pre-trained word embeddings, word clusters (e.g. Brown clusters) or extra resources.}, our model achieves 1.3\% $F_1$ improvement compared to the baseline parser with fully-supervised learning \cite{Zhu:2013up}.
Our model outperforms the state-of-the-art fully-supervised system \cite{Carreras:2008gt,Shindo:2012wk} by 0.6\% $F_1$.
In addition, our fully-supervised model also catches up with many state-of-the-art semi-supervised models \cite{Zhu:2013up,Huang:2009vg,Huang:2010tn,Durrett:2015wy}
%\footnote{Shindo et al.\ \shortcite{Shindo:2012wk} report 92.4\% $F_1$ using ensemble parsing, which is higher than our works.}
by achieving 91.7\% $F_1$ on WSJ test set.  
The size of our model is much smaller than the semi-supervised model of Zhu et al. \shortcite{Zhu:2013up}, which contains rich features from a large automatically parsed corpus.
In contrast, our model is about the same in size compared to the baseline parser.

We carry out Chinese experiments with the same models, and compare the final results with previous related work on the CTB test set.
As shown in Table \ref{chinese final}, our model achieves 2.3\% $F_1$ improvement compared to the state-of-the-art baseline system with fully-supervised learning \cite{Zhu:2013up}, which are by far the best results in the literature.
In addition, our fully-supervised model is also comparable to many state-of-the-art semi-supervised models \cite{Zhu:2013up,Wang:2014bj,Wang:2015gk,Dyer:2016uz} by achieving 85.5\% $F_1$ on the CTB test set. \newcite{Wang:2014bj} and \newcite{Wang:2015gk} do joint POS tagging and parsing.

%%comparison of running time
\subsection{Comparison of Speed}
Table \ref{time} shows the running times of various parsers on test sets on a Intel 2.2 GHz processor with 16G memory.
Our parsers are much faster than the related parser with the same shift-reduce framework \cite{Sagae:2005dw,Sagae:2006jn}.
Compared to the baseline parser, our parser gives significant improvement on accuracies (90.4\% to 91.7\% $F_1$) at the speed of 79.2 sentences per second\footnote{The constituent hierarchy prediction is excluded. The cost of this step is far less than the cost of parsing, and can be eliminated by pipelining the constituent hierarchy prediction and the shift-reduce decoder.}, in contrast to 89.5 sentences per second on the standard WSJ benchmark.
%%%%%%%time
\begin{table}[!tp]
\begin{center}
\renewcommand{\arraystretch}{0.8}
\begin{tabular}{>{\small}l>{\small}c}
\hline
Parser & \#Sent/Second \\
\hline
Ratnaparkhi \shortcite{Ratnaparkhi:1997uj} & Unk \\
Collins \shortcite{collins2003head} & 3.5 \\
Charniak \shortcite{Charniak:2000vi} & 5.7 \\
Sagae and Lavie \shortcite{Sagae:2005dw} & 3.7 \\
Sagae and Lavie \shortcite{Sagae:2006jn} & 2.2 \\
Petrov and Klein \shortcite{Petrov:2007uu} & 6.2 \\
Carreras et al. \shortcite{Carreras:2008gt} & Unk \\
Zhu et al. \shortcite{Zhu:2013up} & 89.5 \\
\hline
This work & 79.2 \\
\hline
\end{tabular}
\end{center}
\caption{\label{time} Comparison of running times on the test set, where the time for loading models is excluded. The running times of related parsers are taken from Zhu et al. \shortcite{Zhu:2013up}.}
\end{table}
%%%%%%%

\section{Errors Analysis}
We conduct error analysis by measuring: parsing accuracies against different phrase types, constituents of different span lengths, and different sentence lengths.

%%different phrase types
\subsection{Phrase Type}
Table \ref{errors analysis on phrase types} shows the accuracies of the baseline and the final parsers with lookahead features on 9 common phrase types.
As the results show, while the parser with lookahead features achieves improvements on all of the frequent phrase types, there are relatively more improvements on constituent \emph{VP}, \emph{S}, \emph{SBAR} and \emph{WHNP}.

The constituent hierarchy predictor has relatively better performance on $s$-type labels for the constituents \emph{VP} and \emph{WHNP}, which are prone to errors by the baseline system.
The constituent hierarchy can give guidance to the constituent parser for tackling the challenges.
Compared to the $s$-type constituent hierarchy, the $e$-type constituent hierarchy is relatively more difficult to predict, particularly for the constituents with long spans such as \emph{VP}, \emph{S} and \emph{SBAR}.
Despite this, the $e$-type constituent hierarchies with relatively low accuracies also benefit prediction of constituents with long spans.

%%%%%%%errors analysis phrase types
\begin{table*}[!tp]
\begin{center}
\renewcommand{\arraystretch}{0.8}
\begin{tabular}{>{\small}l>{\small}l>{\small}c>{\small}c>{\small}c>{\small}c>{\small}c>{\small}c>{\small}c>{\small}c>{\small}c}
\hline
\multicolumn{2}{>{\small}l}{} & NP & VP & S & PP & SBAR & ADVP & ADJP & WHNP & QP \\
\hline
\multicolumn{2}{>{\small}l}{baseline} & 92.06  & 90.63 & 90.28 & 87.93 & 86.93  & 84.83 & 74.12 & 95.03 & 89.32 \\
\multicolumn{2}{>{\small}l}{with lookahead feature} & 93.10 & 92.45 & 91.78 & 88.84 & 88.59 & 85.64 & 74.50 & 96.18 & 89.63\\
\multicolumn{2}{>{\small}l}{improvement} & +1.04 & +1.82 & +1.50 & +0.91 & +1.66 & +0.81 & +0.38 & +1.15 & +0.31 \\
\hline
\hline
\multirow{2}{*}{constituent hierarchy} & $s$-type & 95.18 & 97.51 & 93.37 & 98.01 & 92.14 & 88.94 & 79.88 & 96.18 & 91.70 \\
& $e$-type & 91.98 & 76.82 & 80.72 & 84.80 & 66.82 & 85.01 & 71.16 & 95.13 & 91.02 \\
\hline
\end{tabular}
\end{center}
\caption{\label{errors analysis on phrase types} Comparison between the parsers with lookahead features on different phrases types, with the corresponding constituent hierarchy predictor performances.}
\end{table*}
%%%%%%%
%%different span lengths
\subsection{Span Length}
\begin{figure}[!tp]
\begin{tikzpicture}[scale=1, font=\small]
\begin{axis} [
height = 5cm,
width = 8cm,
xlabel = span length,
ylabel = F$_1$ Score (\%),
xmin=1,
xmax=14,
ymin=83,
ymax=95,
ytick pos=left,
legend style={at={(0.9,0.95)}},
ymajorgrids=true,
grid style=dashed,
y label style={at={(0.05,0.5)}}
]
\addplot[thick] coordinates{
	(1, 92.38)
	(2, 94.11)
	(3, 92.80)
	(4, 90.18)
	(5, 88.44)
	(6, 87.18)
	(7, 87.07)
	(8, 86.09)
	(9, 86.45)
	(10, 85.84)
	(11, 84.64)
	(12, 85.11)
	(13, 86.30)
	(14, 85.00)
%	(15, 85.48)
%	(16, 84.03)
};
\addlegendentry{baseline}
\addplot[dashed] coordinates{
	(1, 93.26)
	(2, 94.75)
	(3, 93.78)
	(4, 91.48)
	(5, 89.77)
	(6, 88.93)
	(7, 88.69)
	(8, 87.65)
	(9, 88.56)
	(10, 87.73)
	(11, 86.60)
	(12, 86.46)
	(13, 89.33)
	(14, 87.62)
};
\addlegendentry{lookahead}
\end{axis}
\end{tikzpicture}
\caption{Comparison with the baseline on spans of different lengths.}
\label{span length} 
\end{figure}
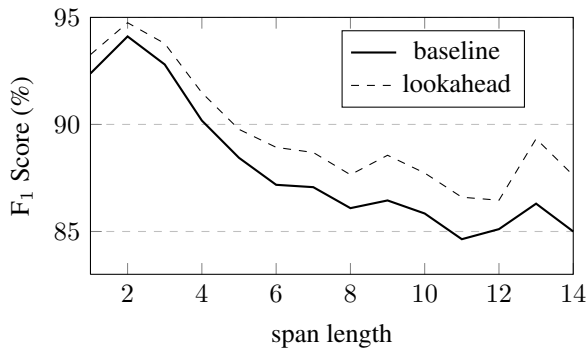

Figure \ref{span length} shows comparison of the two parsers on constituents with different span lengths.
As the results show, lookahead features are helpful on both large spans and small spans, while the performance gap between the two parsers is larger as the size of span increases.
This reflects the usefulness of long-range information captured by the constituent hierarchy predictor and lookahead features.

%%%%%%%span length
%%%%%%%
%%different sentence lengths

\subsection{Sentence Length}
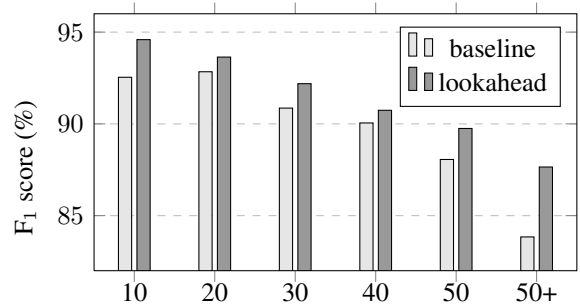
\begin{figure}[!tp]
\begin{center}
\begin{tikzpicture}[scale=1, font=\small]
\begin{axis}[
height = 5cm,
width = 8cm,
    ybar,
    ymin= 82,
    ymax = 96,
    bar width=5pt,
    legend style={at={(0.80,0.95)}, anchor=north},
    ylabel = {F$_1$ score (\%)},
    y label style={at={(0.05,0.4)}},
    x label style={at={(0.5,0.0)}},
    grid style=dashed,
    ymajorgrids=true,
    symbolic x coords={
    10,
    20,
    30,
    40,
    50,
    50+
    },
    xtick=data,
    xticklabel style={
        inner sep=0pt,
    %    anchor=north east,
        rotate=0
    },
    %nodes near coords,
    %nodes near coords align={vertical},
    % every node near coord/.append style={
        %anchor=mid west,
     %   rotate=30
    %},
    ]
    \addplot[fill=gray!20]coordinates {
        (10, 92.54)
	(20, 92.84)
	(30, 90.86)
	(40, 90.05)
	(50, 88.06)
	(50+, 83.84)
     };
     \addplot[fill=gray!80] coordinates {
	(10, 94.59)
	(20, 93.64)
	(30, 92.19)
	(40, 90.74)
	(50, 89.75)
	(50+, 87.65)
     };
     \legend{baseline, lookahead}
\end{axis}
\end{tikzpicture}
\end{center}
\caption{\label{sentence lengths} Comparison with the baseline on sentences of different lengths. Sentences with length [0, 10) fall in the bin 10.}
\end{figure}

Figure \ref{sentence lengths} shows comparison of the two parsers on sentences of different lengths.
As the results show, the parser with lookahead features outperforms the baseline system on both short sentences and long sentences. Also, the performance gap between the two parsers is larger as the length of sentence increases.

The constituent hierarchy predictors generate hierarchical constituents for each input word using global information.
For longer sentences, the predictors yield deeper constituent hierarchies, offering corresponding lookahead features.
As a result, compared to the baseline parser, the performance of the parser with lookahead features decreases more slowly as the length of the sentences increases.
%%%%%%%
%% constituency

%%different numbers of unknown word
%\subsection{Numbers of unknown word}
%Figure \ref{number of unknown word} shows the comparison of the two parsers performance on sentences in different numbers of unknown words.
%As the results show, the fully-supervised parser outperforms the baseline in sentences with unknown word less than 3.
%%%%%%%number of unknown word
%\begin{figure}[!tp]
%\begin{tikzpicture}[scale=1, font=\small]
%\begin{axis} [
%height = 5cm,
%width = 8cm,
%xlabel = the number of unknown word,
%ylabel = F$_1$ Score (\%),
%xmin=0,
%xmax=6,
%ymin=83,
%ymax=93,
%ytick pos=left,
%legend style={at={(0.9,0.95)}},
%ymajorgrids=true,
%grid style=dashed,
%y label style={at={(0.05,0.5)}}
%]
%\addplot[thick] coordinates{
%	(0, 91.27)
%	(1, 89.74)
%	(2, 89.55)
%	(3, 87.30)
%	(4, 88.55)
%	(5, 85.02)
%	(6, 83.33)
%};
%\addlegendentry{baseline}
%\addplot[dashed] coordinates{
%	(0, 92.12)
%	(1, 90.62)
%	(2, 90.35)
%	(3, 89.17)
%	(4, 87.78)
%	(5, 85.42)
%	(6, 90.74)
%};
%\addlegendentry{fully}
%\end{axis}
%\end{tikzpicture}
%\caption{Comparison of parsing performance of the baseline and the parser with fully-supervised on sentences with different numbers of unknown word (up to 6).}
%\label{number of unknown word} 
%\end{figure}
%%%%%%%

%Related work
\section{Related Work}
Our lookahead features are similar in spirit to the pruners of Roark and Hollingshead \shortcite{Roark:2009vd} and Zhang et al. \shortcite{Zhang:2010vo}, which infer the maximum length of constituents that a particular word can start or end.
However, our method is different in three main perspectives.
First, rather than using a CRF with sparse local word window features, a neural network is used for dense global features on the sentence.
Second, not only the size of constituents but also the constituent hierarchy is identified for each word.
Third, the results are added into a transition-based parser as soft features, rather then being used as hard constraints to a chart parser.

Our concepts of \emph{constituent hierarchies} are similar with the work of \emph{supertagging}.
For lexicalized grammars such as Combinatory Categorial Grammar (CCG), Tree-Adjoining Grammar (TAG) and Head-Driven Phrase Structure Grammar (HPSG), each word in the input sentence is assigned one or more super tags, which are used to identify the syntactic role of the word for constraint parsing \cite{Clark:2002uk,Clark:2004ha,Carreras:2008gt,Ninomiya:2006uc,Dridan:2008wt}. 
For a lexicalized grammar, \emph{supertagging} can benefit the parsing in both accuracy and efficiency by offering \emph{almost-parsing} information. 
In particular,  \newcite{Carreras:2008gt} defined the concept \emph{spine} in TAG, which is similar to our \emph{constituent hierarchy}.
However, there are three differences.
First, the \emph{spine} is defined to describe the main syntactic tree structure with a series of unary projections, while \emph{constituent hierarchy} is defined to describe how words can start or end hierarchical constituents (it is possible to be empty if the word cannot start or end constituents).
Second, \emph{spines} are extracted from gold trees and used to prune the search space of parsing as hard constraints.
In contrast, we use constituent hierarchies as soft features.
Third,  \newcite{Carreras:2008gt} use \emph{spines} to prune a chart parsing, while we use \emph{constituent hierarchies} to improve a linear shift-reduce parser.

Under the lexicalized grammar, this \emph{supertagging} can benefit the parsing with more accuracy and efficiency as \emph{almost parsing} \cite{Bangalore:1999th}.
Recently, the works on obtaining the super tags appear.
\newcite{Zhang:2010tg} proposed the efficient methods to obtain super tags for HPSG parsing using dependency information.
\newcite{Xu:2015hw} and \newcite{Vaswani:2016tb} turn to design recursive neural network for \emph{supertagging} for CCG parsing.
In contrast, our models predict the constituent hierarchy instead of single super tag for each word in the input sentence, which are also likely regarded as the member of multiple ordered labels prediction family.

Our constituent hierarchy predictor is also related to sequence-to-sequence learning \cite{Sutskever:2014tya}, which is successful in neural machine translation \cite{Bahdanau:2014vz}.
The neural model encodes the source-side sentence into dense vectors, and then uses them to generate target-side word by word.
There has also been work that directly use sequence-to-sequence model for constituent parsing, which generates bracketed constituency tree given raw sentences \cite{Vinyals:2015uh,Luong:2015uf}.
Compared to Vinyals et al. \shortcite{Vinyals:2015uh}, who predicts a full parser tree from input, our predictors tackle a much simpler task, by predicting the constituent hierarchies of each word separately.
In addition, the outputs of the predictors are used for soft lookahead features in bottom-up parsing, rather than being taken as output structures directly.

By integrating the neural constituent hierarchy predictor, our parser is related to neural network models for parsing, which has given competitive accuracies for both constituency parsing \cite{Dyer:2016uz,Watanabe:2015wa} and dependency parsing \cite{Chen:2014dg,Zhou:2015ti,Dyer:2015tt}.
In particular, our parser is more closely related to neural models that integrates discrete manual features \cite{Socher:2013tn,Durrett:2015wy}. Soccer et al. use neural features to rerank a sparse baseline parser; Durrett and Klein directly integrate sparse features into neural layers in a chart parser.
In contrast, we integrate neural information into sparse features in the form of lookahead features.

There has also been work on lookahead features for parsing.
Tsuruoka et al. \shortcite{Tsuruoka:2011uh} run a baseline parser for a few future steps, and use the output actions to guide the current action.
In contrast to their model, our model leverages full sentential information, yet is significantly faster.

Existing work on investigated efficiency of parsing without loss of accuracy, which is required by real time applications, such as web parser, processing massive amounts of textual data.
\newcite{Zhang:2010vo} introduced a chart pruner to accelerate a CCG parser.
\newcite{Kummerfeld:2010wo} proposed novel self-training method focusing on increasing the speed of a CCG parser rather than its accuracy.
%For some offline tasks such as , speed may be not as crucial.

%conclusion
\section{Conclusion}
We proposed a novel constituent hierarchy predictor based on recurrent neural networks, aiming to capture global sentential information.
The resulting constituent hierarchies are fed to a baseline shift-reduce parser as lookahead features, addressing the limitation of shift-reduce parsers in not leveraging right-hand side syntax for local decisions, yet maintaining the same model size and speed.
The resulting fully-supervised parser outperforms the state-of-the-art baseline parser by achieving 91.7\% $F_1$ on standard WSJ evaluation and 85.5\% $F_1$ on standard CTB evaluation.

\section*{Acknowledgments}
We thank the anonymous reviewers for their detailed and constructed comments.
This work is supported by T2MOE 201301 of Singapore M-O-E.
Yue Zhang is the corresponding author.

%%%%
%%%%

\bibliography{acl2016}
\bibliographystyle{acl2012}

\end{document}